\documentclass[conference,a4paper]{IEEEtran}
\IEEEoverridecommandlockouts

\usepackage{eso-pic}
\newcommand{\CopyrightNotice}{\AddToShipoutPictureFG*{%
  \AtPageLowerLeft{\raisebox{10mm}{\hspace*{\dimexpr(\paperwidth-\textwidth)/2\relax}%
  \parbox{\textwidth}{\footnotesize \copyright~2026 IEEE. Personal use of this material is permitted. Permission from IEEE must be obtained for all other uses, in any current or future media, including reprinting/republishing this material for advertising or promotional purposes, creating new collective works, for resale or redistribution to servers or lists, or reuse of any copyrighted component of this work in other works.}}}}}
\usepackage{cite}
\usepackage{amsmath,amssymb,amsfonts}
\usepackage{graphicx}
\usepackage{textcomp}
\usepackage{xcolor}
\usepackage{booktabs}
\usepackage{tabularx}
\usepackage{colortbl}
\usepackage{url}
\usepackage{placeins}
\usepackage{multirow}
\usepackage{longtable}
\usepackage{enumitem}
\setlist{itemsep=1pt,topsep=2pt,parsep=0pt}
\usepackage{tikz}
\usetikzlibrary{shapes.geometric, arrows.meta, positioning, fit, backgrounds, calc}

\usepackage{dblfloatfix}
\usepackage{microtype}
\begin{document}

\title{A Simulation Platform for AUV Fault Recovery: Exploring LLM-Based Diagnostic Strategies}

\author{
\IEEEauthorblockN{Khalid Halba,\quad Kylie Cooper,\quad James G. Bellingham}
\IEEEauthorblockA{\textit{Exploration Robotics Laboratory},
\textit{Johns Hopkins Institute for Assured Autonomy}, Baltimore, MD, USA\\
khalba1@jhu.edu \quad kcoope68@jh.edu \quad jbellingham@jhu.edu}
}

\maketitle
\CopyrightNotice

\begin{abstract}
Autonomous underwater vehicles (AUVs) operating beyond reliable communications must recover from failures without human intervention. We investigate an architecture in which conventional deterministic layered control autonomy manages normal operations, while an invokable large language model (LLM) serves as a diagnostic and recovery planner when onboard anomaly detection identifies performance outside expected limits. Because language models are stochastic, rigorous evaluation requires ensemble testing rather than individual demonstrations. We present a closed-loop simulation architecture that couples real-time C vehicle software with a higher-level orchestration layer for physics-based fault injection, structured prompting, language-model interaction, mission file generation, validation, execution, and LLM-judge scoring. The framework, which we call SPAR (Simulation Platform for AUV Recovery), supports evaluation across fault realizations, prompt structures, reasoning models, and mission conditions. We vary these for a mass-shift fault over 480 SPAR trials, evaluating a frontier model and three off-the-shelf locally deployable LLMs.  Model choice dominates diagnosis: the frontier model places the CG-shift mechanism in its top three hypotheses in 85--90\% of trials, versus 60--78\% for the best local model. Reasoning analysis indicates that local-model success is associated with following the complete diagnostic procedure, whereas weaker models often commit prematurely to elevator failure even though the actuator tracks its command. Diagnosis and operational decision performance do not appear to be coupled in this dataset. The contributions are an architecture extending unanticipated-fault recovery from detection to mitigation and an ensemble methodology for evaluating LLM-assisted mission management on low-power AUVs.
\end{abstract}


\section{Introduction and Related Work}
\label{sec:intro}

\subsection{Context and Contributions}
Long-duration AUV missions are normally executed by deterministic layered control autonomy that manages guidance, control, behaviors, and mission sequencing. This architecture is appropriate for nominal operation and for anticipated faults with predefined responses. The unresolved case is mission management after an unanticipated failure. The vehicle observes out-of-bounds performance, but the cause and appropriate operational decision are not encoded in the existing layered control autonomy. This paper examines the use of a language model as an invokable diagnostic and recovery planner for that case. The LLM does not replace the real-time layered control autonomy. The vehicle continues to operate under its conventional layered control until an anomaly detector identifies performance outside expected limits. The vehicle then assembles a structured query containing mission state, vehicle state, sensor history, actuator status, and available recovery actions. The language model returns a diagnosis and candidate recovery plan, which is validated for format before execution.

This architecture is motivated by the deployment constraints of AUVs operating with limited or unavailable communications, such as under-ice missions. It also reflects power and computational constraints as even small language models (SLMs) on edge hardware such as the NVIDIA Jetson~\cite{b12} can more than double the idle (hotel-load) power of a vehicle as power-frugal as the Tethys long-range AUV (LRAUV)~\cite{b30}.
Consequently, real-time control remains in C code suitable for low-power processors. The language model is invoked only after anomaly detection. In the intended onboard implementation, the planner would be a small local model rather than a frontier cloud model. Onboard inference is required by limited-communications operation but faces severe edge-hardware constraints~\cite{b25}.

Although retrieval augmentation~\cite{b28}, agentic tool use~\cite{b29}, and fine-tuning~\cite{b26} may improve performance, this study establishes a baseline using off-the-shelf LLMs with a single response per trial and no iterative re-prompting.

We implement this evaluation architecture building on the MIT Sea Grant Odyssey~II simulator, with real-time C vehicle dynamics and mission logic coupled to a Python orchestration layer for fault injection, structured prompting, model invocation, mission file generation, validation, execution, and scoring. The original C-simulator has been supplemented with parametric models of vehicle subsystems for injecting realistic failures. The result is a closed-loop test facility for reasoning-enabled AUV autonomy. We demonstrate the system on a failure that occurred in an MBARI (Monterey Bay Aquarium Research Institute) LRAUV mission~\cite{b13}, a center-of-gravity mass-shift fault, and we evaluate three levels of diagnostic framing, reporting ensemble results over 480 SPAR trials. The contribution is an experimental architecture for testing language-model-assisted AUV mission recovery, together with initial measurements of how diagnostic accuracy and operational decisions vary with model class, prompt information, fault magnitude, and mission phase.

\subsection{Related Work}
The work is organized around two related AUV platforms. The MIT Sea Grant Odyssey~II~\cite{b1,b2} was an early ($140$--$200$~kg) torpedo-type AUV with an extensive field history, including Arctic, deep-ocean, coastal, and multi-vehicle deployments; its vehicle and simulator code base provides the layered control autonomy used in SPAR. The MBARI Tethys-class LRAUV~\cite{b30} (similar weight class) provides the operational context motivating the study: it is an active, long-endurance scientific platform, with more than $36{,}000$ offshore hours accumulated across the fleet. Its operational history provides field-grounded failure cases from which SPAR scenarios can be constructed, including the mass-shift event examined here. We note that the Tethys software evolved from the Odyssey~II code, so the two share a common heritage.

Undersea robotics has a broad simulator ecosystem, ranging from vehicle-specific software-in-the-loop testbeds to general-purpose robotics and game-engine platforms extended with hydrodynamics, actuators, and underwater sensors. HoloOcean builds on Unreal Engine to provide high-fidelity environmental and sensor simulation~\cite{b31}, while UUV Simulator extends Gazebo with underwater vehicle, actuator, and sensor models~\cite{b32}. Other platforms, including Stonefish, DAVE, and the AUV Workbench, provide related capabilities in physics-based simulation, mission rehearsal, and vehicle-software integration~\cite{b34,b35,b36}. SPAR shares this general structure but is organized specifically for fault modeling, repeatable fault injection, and closed-loop operational decision evaluation against proven vehicle code.

Classical AUV fault-diagnosis methods~\cite{b19} predominantly target predefined actuator, sensor, and component faults using model-based residuals, thresholds, observers, or trained classifiers.  Coverage of unknown or derived faults remains limited. Raanan et al.~\cite{b4,b13} advanced task-level monitoring by detecting unanticipated departures
from nominal vehicle behavior using online topic models and a real-time vertical-plane anomaly detector. These methods establish that the vehicle is behaving abnormally, but do not determine the underlying physical cause or generate a mission-level operational decision. This work addresses those subsequent diagnostic and mitigation steps.

LLMs have been used to translate natural-language goals into executable plans for ground and aerial robots~\cite{b5,b18}, but their application to AUV fault recovery remains limited. The closest concurrent work~\cite{b21} considers operator-facing anomaly diagnosis, while earlier planning studies generally assume fault-free execution. We
instead use the LLM as a text-based diagnostic and recovery planner that interprets structured anomaly telemetry, proposes a new mission file, and passes that mission to a deterministic validator before closed-loop execution. Because LLM physical reasoning remains imperfect~\cite{b16}, SPAR evaluates this process statistically under
controlled vehicle faults.

\begin{figure}[t]
  \centering
  \includegraphics[width=\columnwidth]{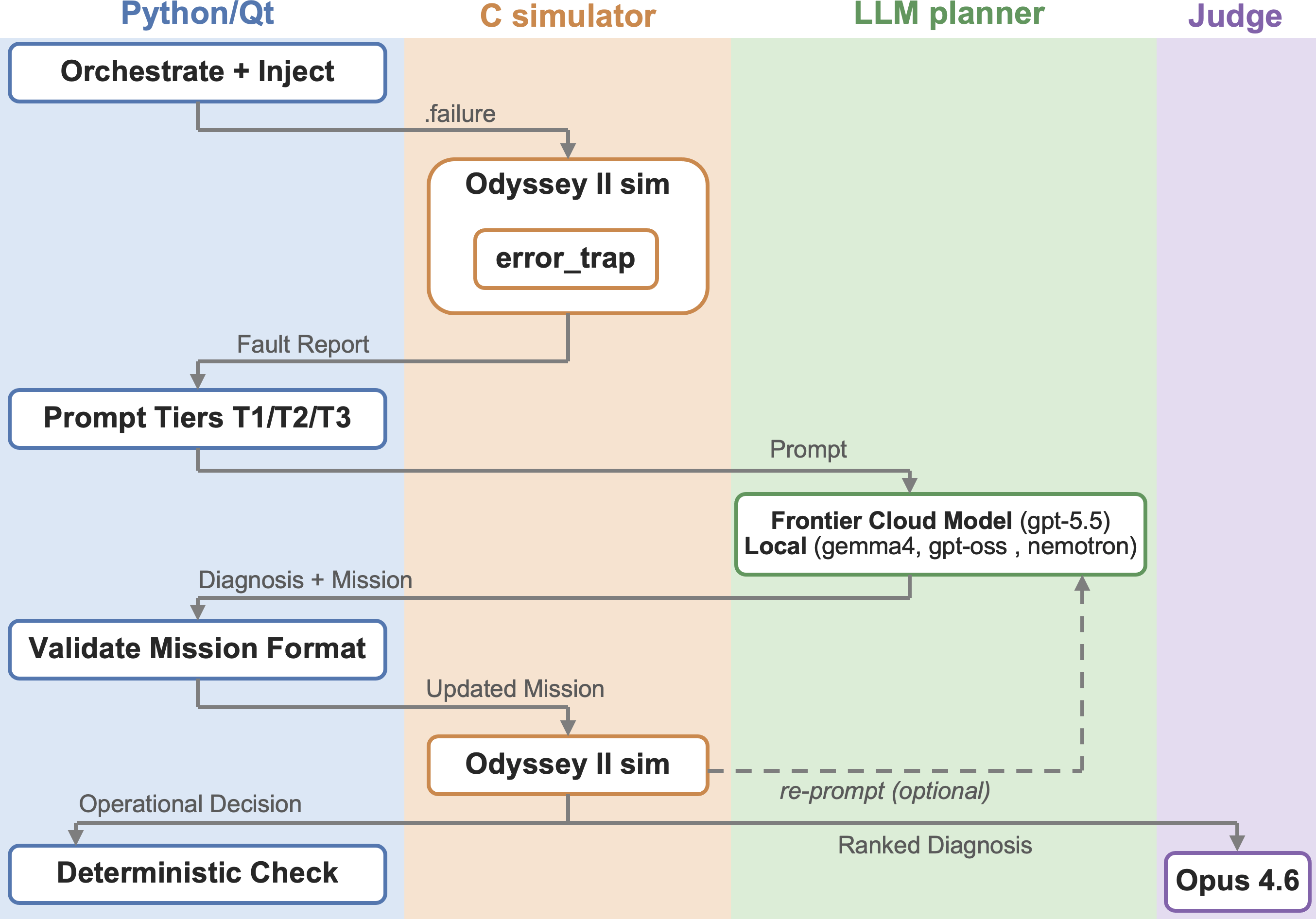}
  \caption{The SPAR pipeline by subsystem lane, top-down from injection to scoring. The dashed arrow marks the optional re-prompt path, unused here.}
  \label{fig:arch}
\end{figure}

\section{Software Architecture}
\label{sec:method}
\subsection{Software Components}
\label{sec:platform_arch}

\begin{figure*}[t]
\centering
\includegraphics[width=\textwidth]{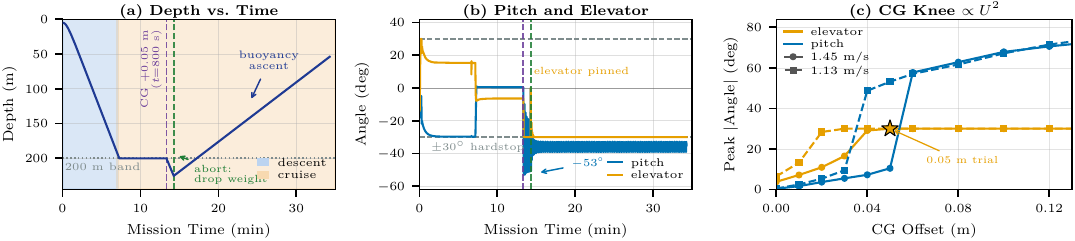}
\caption{\textbf{(a)} CG-shift trial (\texttt{gemma4:12b}, Tier~1, $0.05$~m at $t{=}800$~s), depth vs.\ mission time. \textbf{(b)} The shift pins the elevator at $-30^\circ$, so buoyancy (not control) recovers. \textbf{(c)} Peak elevator and pitch vs.\ CG offset, two speeds. The knee moves out ($0.024{\to}0.040$~m), the $0.05$~m trial (star) past it.}
\label{fig:ocg_trial}
\end{figure*}
SPAR is organized as four software components, shown as lanes in Fig.~\ref{fig:arch}: a Python/Qt orchestration layer, the C vehicle simulator, the language-model planner under test, and a language-model judge. The components exchange a small set of files and data structures that close the loop from fault injection to a scored operational decision. The orchestration layer writes a fault descriptor that the simulator reads. The simulator returns a per-control-cycle (5 Hz) sensor table and, once its detector fires, an anomaly flag. The orchestration layer then assembles a prompt for the planner, and the planner returns a diagnosis together with a mission file. The validated mission is executed, and the judge scores the resulting diagnosis and operational decision.

\paragraph{Orchestration Layer (Python/Qt)}
\label{sec:comp_orch}
The orchestration layer is modular and coordinates the other three components. It writes the fault descriptor that seeds injection, launches the vehicle binary, ingests the sensor table and the anomaly flag, assembles the prompt (Section~\ref{sec:prompts}), validates the returned mission for syntax and safety, and routes the operational decision outcome to the judge. The layer runs no learned fault classifier, so all diagnostic reasoning is left to the planner.

\paragraph{Vehicle Simulator (C)}
\label{sec:comp_sim}
The vehicle simulator is a modified Odyssey~II binary that integrates six-degree-of-freedom hydrodynamics at $50$~Hz~\cite{b6}. It reads the fault descriptor and injects the specified fault into the running vehicle, recording the vehicle's sensed variables to the sensor table. An error trapping module watches four sensed channels through three filtering stages, namely deadband, persistence, and verification, and raises an anomaly flag when a channel breaches its threshold. The simulator also exposes deterministic behaviors that a mission can invoke, and it executes the planner's mission file to produce the resulting trajectory.

\paragraph{Language-Model Planner}
\label{sec:comp_planner}
The planner is the model under evaluation, either the frontier \texttt{gpt-5.5} or a locally deployable model such as \texttt{gemma4}, \texttt{gpt-oss}, or \texttt{nemotron}. The planner is a swappable component, so any local or cloud model reachable through the same text interface can be used in its place, the four here being representative. It receives the assembled prompt and returns a single response that carries a diagnosis, an executable mission in Odyssey~II's 1998 mission-file format, and, when the model exposes it, an intermediate reasoning trace. Local models run under Ollama on a single $24$~GB RTX~3090. Median wall-clock LLM time per trial, including validation retries, ranged from about $45$~s for \texttt{gemma4} to about $250$~s for the larger local models (\texttt{gpt-oss} ${\sim}251$~s, \texttt{nemotron} ${\sim}240$~s)~\cite{b11}.

\paragraph{Language-Model Judge}
\label{sec:comp_judge}
The judge is a separate frontier model, \texttt{claude-opus-4-6}, drawn from a fourth vendor so no model grades itself. It reads the planner's diagnosis and the operational decision outcome and returns a diagnostic score~\cite{b14}. The continue-or-abort decision is checked separately by deterministic code against the fault physics rather than by the judge.

\subsection{LLM Prompt}
\label{sec:prompts}
The LLM prompt is assembled per run from static and dynamic sections. The static sections hold a fixed engineering reference over eight physical domains (hydrostatic pressure and buoyancy, drag and power, propulsion, propeller thrust, static stability, attitude--depth kinematics, depth control, and the ocean environment), the behavior catalog, and the Odyssey~II mission-file format. No subsystem is named as the source, and competing mechanisms are given comparable detail. The dynamic sections, supplied by the simulator and orchestration layer, hold the fault report, vehicle and actuator status, power state, and time-history telemetry. The fault report names only the detected symptom, never the injected fault or its magnitude. The prompt shows a subset of the $117$ sensed variables sufficient to characterize the anomaly. Internal quantities such as mass distribution, net buoyancy, and moment balances are unsensed and inferred from their effects, as on a real vehicle.

\subsection{Workflow}
\label{sec:workflow}
The operator picks a vehicle, here the Odyssey~II simulator binary; a site, either a location with bathymetry or open water; and a mission such as a descent-and-cruise profile. Fault injection allows any phase, a range of magnitudes, and several faults per mission. The operator then sets the run mode (instant, real time, or overnight sweep), the planner models, which do both diagnosis and operational decision, and a judge that rates whether the diagnosis identifies the fault and whether the continue-or-abort choice is sound.

\begin{table*}[!t]
\centering
\caption{Physical ground truth for the four CG-shift experiment cases.
The authority ratio is
$\chi=\Delta x/\Delta x_{\mathrm{knee}}(U)$; $\chi<1$ indicates that
the disturbance remains within measured elevator authority.}
\label{tab:cg_cases}

\footnotesize
\setlength{\tabcolsep}{3pt}
\renewcommand{\arraystretch}{1.00}

\begin{tabularx}{\textwidth}{
@{}
l
c
c
c
c
>{\raggedright\arraybackslash}X
c
@{}}
\toprule
Flight phase &
\shortstack{$U$\\(m/s)} &
\shortstack{$\Delta x_{\mathrm{knee}}$\\(m)} &
\shortstack{$\Delta x$\\(m)} &
$\chi$ &
Expected vehicle response &
\shortstack{Correct\\action} \\
\midrule

Descent &
1.45 &
0.040 &
0.005 &
0.13 &
Small added nose-down moment during commanded descent.
Sufficient elevator authority remains.
The fault is trimmable.
The nominal descent partly masks the depth signature. &
\textbf{Continue} \\

Descent &
1.45 &
0.040 &
0.050 &
1.25 &
The added moment exceeds elevator authority.
The elevator saturates.
Pitch becomes more nose-down than commanded.
Depth increases faster than the nominal descent. &
\textbf{Abort} \\

\addlinespace[2pt]

Cruise &
1.13 &
0.024 &
0.005 &
0.21 &
Small persistent nose-down trim disturbance during depth hold.
Sufficient elevator authority remains.
Depth control remains feasible. &
\textbf{Continue} \\

Cruise &
1.13 &
0.024 &
0.050 &
2.08 &
The added moment substantially exceeds elevator authority.
The elevator saturates.
Nose-down pitch persists.
Depth leaves the commanded 200~m band. &
\textbf{Abort} \\

\bottomrule
\end{tabularx}
\end{table*}

\section{Experiment Design}
\label{sec:experiment}

\subsection{Mass-Shift Fault}
\label{sec:f4_fault_model}

We simulate an uncommanded forward shift of the vehicle center of
gravity (CG), representing a displaced internal mass or material
attached during a collision. The scenario is motivated by Tethys-class
LRAUV incidents in which a battery mass shifted forward, producing a
nose-down attitude beyond $-30^\circ$ with the stern plane saturated,
and a later mass-shifter fault produced a slow trim drift
of $0.06^\circ$/hour~\cite{b13}.

For nose-up pitch $\theta$, forward CG displacement $\Delta x$, vehicle
weight $W$, and nominal vertical separation $BG$ between the center of
buoyancy and CG, the hydrostatic pitch moment is
\begin{equation}
M_h(\theta) =
-W\left(BG\sin\theta+\Delta x\cos\theta\right).
\end{equation}
With $\Delta x=0$, the restoring moment gives a stable equilibrium at
$\theta=0$. A forward shift adds a nose-down moment and moves the
uncontrolled equilibrium to
$\theta_{\mathrm{eq}}=-\tan^{-1}(\Delta x/BG)$.

The resulting deviation from commanded depth is caused by the altered vehicle
attitude, not by a direct downward force. Odyssey~II changes depth by pitching and moving along its longitudinal
axis. With depth positive downward,
\begin{equation}
\dot{z} \simeq -U\sin\theta + w_b ,
\end{equation}
where $U$ is forward speed and $w_b$ is the vertical velocity associated
with net buoyancy. A persistent nose-down pitch therefore produces a
positive depth rate. If the elevator cannot reject the CG-induced moment, the pitch error remains and the vehicle diverges from its commanded depth (Fig.~\ref{fig:ocg_trial}(a)).

Elevator authority increases approximately with $U^2$ but is limited by
fin stall and the $\pm30^\circ$ mechanical stop (Fig.~\ref{fig:ocg_trial}(b)). Consequently, the
effect of a given CG shift depends on both fault magnitude and flight
phase. Table~\ref{tab:cg_cases} summarizes the four experiment cases
using the measured saturation knees (Fig.~\ref{fig:ocg_trial}(c)) at the descent and cruise speeds. The $0.005$~m shift remains within elevator authority in both phases and is therefore assigned a continue response. The $0.05$~m shift exceeds available authority in both phases, producing sustained nose-down pitch and depth divergence, and is assigned a drop-weight abort. During descent, the fault signature is partly masked by the commanded nose-down attitude and increasing depth; during cruise, the same departure is more conspicuous against nominal level flight and constant depth.

\subsection{Prompt Design}

The three prompt tiers were designed to evaluate how progressively reduced engineering guidance affects an LLM's ability to diagnose an AUV anomaly and generate an appropriate mission file while keeping the operational task unchanged. As described in Section~\ref{sec:prompts}, each prompt includes the mission objectives and acceptable operational risk, vehicle and actuator status, power state, time-history telemetry, an engineering reference, the available behaviors, and the required Odyssey~II mission-file format. Using this information, the model is instructed to identify and rank candidate physical failure modes, determine whether sufficient control authority remains to continue the mission safely or whether an immediate abort is required, and generate a valid mission file implementing that decision.

The three prompt tiers differ only in the amount of engineering guidance provided to support that reasoning. Tier~1 provides the most comprehensive engineering descriptions, subsystem-specific context, safety guidance, and governing equations. Tier~2 retains the same operational information while removing the governing equations and most subsystem-specific context. Tier~3 provides only the essential engineering reference and operational constraints, requiring the model to rely more heavily on its own marine engineering knowledge.

The prompt was constructed to avoid preferentially favoring any diagnostic hypothesis. The engineering domains listed in Section~\ref{sec:prompts} are presented independently and with comparable detail. No subsystem is identified as the source of the anomaly, and the telemetry is provided without diagnostic interpretation. Rather than emphasizing static stability or trim calculations, the prompt requires the model to identify and rank candidate mechanisms across the engineering reference before selecting an operational decision. The resulting design therefore evaluates diagnostic reasoning from the observed telemetry.

\subsection{Experiment Case Matrix}

The experiment evaluates four LLMs: one frontier model, \texttt{gpt-5.5}, and three local models, \texttt{gemma4:12b}, \texttt{gpt-oss:20b}, and \texttt{nemotron-nano-12b-v2}. Each model is tested with three prompt tiers, two CG-shift magnitudes ($0.005$ and $0.05$~m), and two injection phases: descent ($t{=}300$~s) and cruise ($t{=}800$~s). Each condition is repeated ten times, giving $480$ SPAR trials and $n{=}40$ trials for each model--tier pair. The same mission is used for every trial: initialize at 5~m, descend to a 200~m band, cruise ${\sim}1$~km.

\subsection{Scoring Criteria}

Each trial produces a ranked diagnosis, an operational decision, an executable Odyssey~II mission file, and, when available, a reasoning trace. Because fault troubleshooting often requires maintaining multiple plausible hypotheses, diagnostic credit does not require the correct mechanism to be ranked first. A diagnosis is scored as correct when a CG shift, mass shift, or physically equivalent mechanism appears within the model's top three hypotheses.  A trial is scored as diagnostically incorrect when the mechanism does not appear within the top three. Rank-one performance is reported separately.

The operational decision is scored independently from the diagnosis using the simulated controllability results. Continuing is correct for the manageable $0.005$~m shift, while aborting is correct for the $0.05$~m shift. Generated mission files are checked for valid Odyssey~II syntax before execution. Reasoning traces support qualitative analysis but are not scored separately; \texttt{gemma4:12b} is evaluated from its final response because reasoning mode was disabled due to unstable non-terminating outputs.

\begin{figure*}[t]
  \centering
  \includegraphics[width=\textwidth]{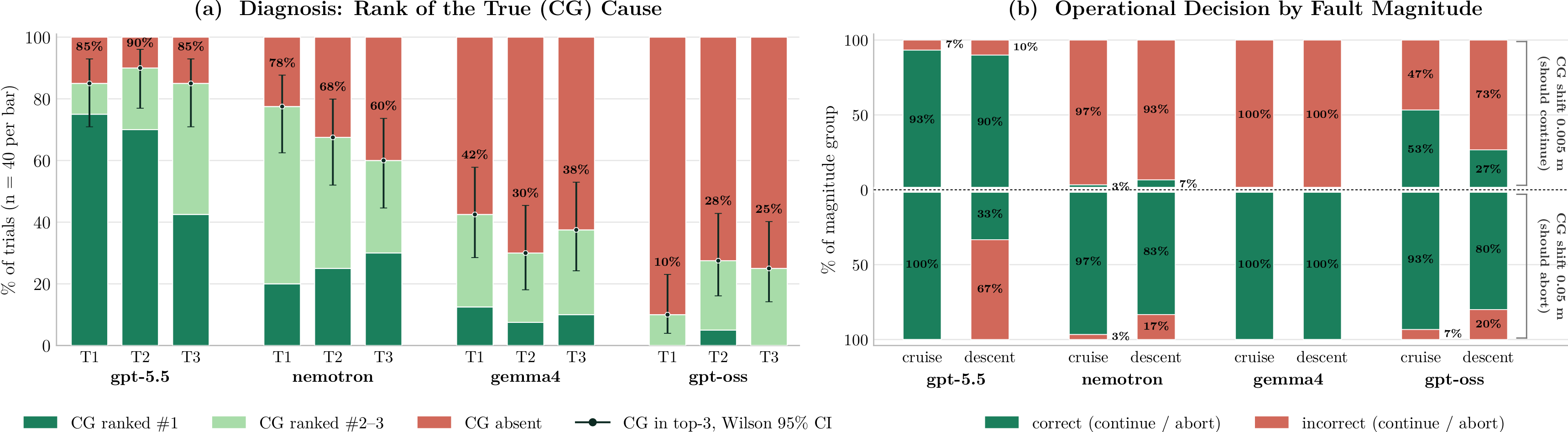}
    \caption{\textbf{(a)} Diagnosis ranking of the true cause in each model's top-3 diagnosis, per model$\,\times\,$tier; labels give the CG-in-top-3 rate and its Wilson 95\% CI. \textbf{(b)} Operational decision by magnitude, per model$\,\times\,$phase.}

  \label{fig:committed_forest}
\end{figure*}

\section{Experimental Results and Discussion}
\label{sec:results}
\subsection{Results}
\subsubsection{Diagnostic Performance} We report Wilson $95\%$ confidence intervals for the top-three diagnostic rates. Fig.~\ref{fig:committed_forest}(a) shows substantial differences among models. \texttt{gpt-5.5} placed the injected CG shift within its top three diagnoses in 85\%, 90\%, and 85\% of Tier~1, Tier~2, and Tier~3 trials, respectively. \texttt{nemotron-nano-12b-v2} had the highest rates among the local models at 78\%, 68\%, and 60\%. \texttt{gemma4:12b} reached 42\%, 30\%, and 38\%, while \texttt{gpt-oss:20b} reached 10\%, 28\%, and 25\%. The confidence intervals overlap across tiers within each model, so the present sample does not resolve a prompt-tier effect. Rank-one performance was lower. The true cause was ranked first by \texttt{gpt-5.5} in 75\%, 70\%, and 43\% of trials. Rank-one rates ranged from 20--30\% for \texttt{nemotron-nano-12b-v2}, 8--13\% for \texttt{gemma4:12b}, and 0--5\% for \texttt{gpt-oss:20b}.

\subsubsection{Operational Decision}
Fig.~\ref{fig:committed_forest}(b) reports continue-or-abort accuracy by fault magnitude and mission phase. Each result combines 30 trials across the three prompt tiers. For the manageable $0.005$~m shift, \texttt{gpt-5.5} correctly continued in 93\% of cruise trials and 90\% of descent trials. \texttt{gpt-oss:20b} reached 53\% and 27\%, while \texttt{nemotron-nano-12b-v2} reached 3\% and 7\%. \texttt{gemma4:12b} aborted every manageable-fault trial. For the $0.05$~m shift, \texttt{gemma4:12b} correctly aborted every trial. \texttt{nemotron-nano-12b-v2} reached 97\% in cruise and 83\% in descent, while \texttt{gpt-oss:20b} reached 93\% and 80\%. \texttt{gpt-5.5} reached 100\% in cruise but 33\% in descent. Across the four conditions, \texttt{gpt-oss:20b} had the highest decision accuracy among the local models.

\subsubsection{LLM Reasoning Traces}
Table~\ref{tab:prompt_processing} summarizes how the local models used the engineering reference after being instructed to examine all eight domains. The \texttt{nemotron} and \texttt{gpt-oss} values are derived from reasoning traces, while the \texttt{gemma4} values are inferred from final responses and indicate only which sections appeared in the reported answer. \texttt{nemotron} used static stability substantively in 94\% of trials, compared with 62\% for \texttt{gpt-oss}. This coincided with the highest local-model diagnostic performance, with the CG shift appearing among \texttt{nemotron}'s top three diagnoses in 60--78\% of trials across tiers. \texttt{gpt-oss} analyzed the continue-or-abort risk trade-off substantively in 90\% of trials, compared with 15\% for \texttt{nemotron}. It also achieved the highest overall decision accuracy among the local models. In its final responses, \texttt{gemma4} most often referenced depth control and attitude--depth kinematics.

\subsection{Discussion}
\label{sec:discussion}

The ensemble dataset produced by SPAR provides insight into model performance under the tested conditions. The local SLMs exhibited distinct strengths and failure modes. \texttt{nemotron-nano-12b-v2} produced the strongest diagnostic results among the local models but remained conservative in its operational decisions. \texttt{gpt-oss:20b} identified the CG shift less often, yet made better decisions for the manageable fault. \texttt{gemma4:12b} always selected an abort and often interpreted the telemetry incorrectly. In several final responses, it inferred an elevator tracking error even though the measured elevator followed its command. These results indicate that telemetry interpretation, fault diagnosis, and operational decision making are separate failure modes.

\begin{table}[t!]
\centering
\footnotesize
\setlength{\tabcolsep}{4pt}
\renewcommand{\arraystretch}{1.06}
\caption{Prompt-section engagement percentage by model, with substantive-reasoning percentages in parentheses.
$^{*}$\texttt{gemma4} values are inferred from final responses, not reasoning traces.}
\label{tab:prompt_processing}
\begin{tabularx}{\columnwidth}{@{}Xccc@{}}
\toprule
Prompt section & \texttt{nemotron} & \texttt{gpt-oss} & \texttt{gemma4}$^{*}$ \\
\midrule
Hydrostatic pressure \& buoyancy & 99 (96) & 99 (96) & 53 (28) \\
Drag \& power                    & 84 (73) & 91 (43) & 39 (13) \\
Propulsion                       & 98 (98) & 94 (82) & 25 (23) \\
Propeller thrust                 & 60 (39) & 54 (15) & \phantom{0}4 (\phantom{0}0) \\
Static stability                 & 97 (94) & 81 (62) & 37 (14) \\
Attitude--depth kinematics       & 98 (79) & 99 (92) & 83 (62) \\
Depth control                    & 96 (82) & 100 (100) & 100 (100) \\
Ocean environment                & 60 (58) & 55 (12) & 40 (34) \\
\midrule
Analyze mission risk  & 52 (15) & 100 (90) & 87 (14) \\
\bottomrule
\end{tabularx}
\end{table}

Reasoning traces provide useful insight into model process, but are not uniformly available across models. The \texttt{gpt-5.5} interface did not expose a reasoning trace. Reasoning mode was disabled for \texttt{gemma4:12b} because it could produce non-terminating sequences. Its values in Table~\ref{tab:prompt_processing} are therefore inferred from final responses and may omit prompt sections that were considered but not reported. For models with traces, the records provide evidence of instruction adherence, treatment of contradictory data, and the integrity of the diagnostic process. This distinction is important in safety-critical autonomy. A correct action without a supported diagnostic process provides less assurance that the model will respond correctly under a different fault or mission condition.

Table~\ref{tab:prompt_processing} shows that \texttt{nemotron-nano-12b-v2} consistently followed the requested diagnostic procedure. It examined multiple subsystems, retained competing hypotheses, and used the static-stability material more often than the other local models. This behavior coincided with its higher diagnostic accuracy. The result suggests that diagnosis depends on both relevant engineering context and a structured procedure for testing candidate mechanisms against telemetry. \texttt{gpt-oss:20b} used equations and quantitative checks most often, but frequently did so after committing to an elevator fault. Additional mathematical analysis did not improve diagnosis when the available evidence was not used to challenge the initial hypothesis.

Operational decision making followed a different pattern. Only \texttt{gpt-oss:20b} consistently performed a substantive continue-or-abort risk analysis. It was also the local model most likely to continue after the manageable $0.005$~m shift. The other local models generally selected abort when the evidence remained uncertain. The severe $0.05$~m shift presented a clear loss of control authority and was easier to act on. The smaller mass shift required recognition of remaining authority and increased operational risk. Fault detection and response also depended strongly on mission phase because the immediate consequences differed between descent and level flight.

The selective use of prompt sections and the weak, non-monotonic response to prompt tier show that providing additional engineering material does not ensure that an SLM will use it effectively. The present experiment does not separate the effects of prompt length, section order, procedural requirements, and model capability. Evaluating prompt composition will require larger experimental runs and controlled ablation studies. The results motivate shorter staged prompts, retrieval of engineering material relevant to the observed anomaly, agentic use of diagnostic tools, and task-specific fine-tuning.

Ultimately, the LLM diagnostic and recovery planner and the layered control autonomy must be engineered as an integrated system. For example, the layered control autonomy should be capable of putting the vehicle in a safe state while the onboard model generates a mitigation plan.  It should also compute and expose derived indicators such as command--response consistency, actuator saturation, remaining control authority, trim state, and energy margin, and provide deterministic checks on critical decisions.  We have focused on the LLM implementation in this paper, but the process exposes lessons for the entire AUV system.

\section{Conclusion}
\label{sec:future}
We presented SPAR as a closed-loop framework for evaluating LLM-assisted diagnosis and recovery from unanticipated AUV faults. The ensemble results show that repeated trials are essential: individual demonstrations do not reveal stochastic variability, recurring failure modes, or systematic differences in how models use evidence. They also show that instruction adherence and evidence-consistent reasoning cannot be assumed. Onboard models will therefore require structured diagnostic procedures, explicit checks of evidence for and against competing hypotheses, selective retrieval of relevant engineering information, and deterministic verification of critical decisions. Diagnostic accuracy and operational decision making should be evaluated separately, since a model may select an appropriate operational decision without correctly identifying the underlying fault.  The analysis provides guidance beyond model selection. SPAR assists in closing the overall AUV system development loop by allowing changes to the model, prompt architecture, deterministic software, sensing, and vehicle design to be evaluated under repeatable physical conditions.

\section*{Acknowledgment}
This work was supported by the Office of Naval Research under Contract N00024-22-D-6404 and by startup funding provided through the Bloomberg Distinguished Professorships Program at Johns Hopkins University.


\begin{thebibliography}{00}
\bibitem{b12} NVIDIA, ``Jetson modules,'' NVIDIA Developer, 2025. [Online]. Available: https://developer.nvidia.com/embedded/jetson-modules. [Accessed: Jul.~12, 2026].
\bibitem{b30} B.~W. Hobson \textit{et al.}, ``Tethys-class long range AUVs -- extending the endurance of propeller-driven cruising AUVs from days to weeks,'' in \textit{Proc. IEEE/OES Autonomous Underwater Vehicles (AUV)}, Southampton, U.K., Sep. 2012, pp.~1--8.
\bibitem{b25} G.~Cai, R.~Tian, L.~Yang, Y.~Jia, L.~Li, and J.~Wang, ``Efficient inference for edge large language models: A survey,'' \textit{Tsinghua Sci. Technol.}, vol.~31, no.~3, pp.~1365--1380, Jun. 2026.
\bibitem{b28} K.~Shuster \textit{et al.}, ``Retrieval augmentation reduces hallucination in conversation,'' in \textit{Findings Assoc. Comput. Linguistics: EMNLP 2021}, 2021, pp.~3784--3803.
\bibitem{b29} S.~Yao \textit{et al.}, ``ReAct: Synergizing reasoning and acting in language models,'' in \textit{Proc. Int. Conf. Learn. Represent. (ICLR)}, 2023.
\bibitem{b26} B.~Raimondi, S.~Giallorenzo, and M.~Gabbrielli, ``Affordably fine-tuned LLMs provide better answers to course-specific MCQs,'' in \textit{Proc. 40th ACM/SIGAPP Symp. Appl. Comput. (SAC)}, 2025, pp.~32--39.
\bibitem{b13} B.~Y. Raanan \textit{et al.}, ``A real-time vertical plane flight anomaly detection system for a long range autonomous underwater vehicle,'' in \textit{Proc. OCEANS 2015 -- MTS/IEEE Washington}, Oct. 2015, pp.~1--6.
\bibitem{b1} J.~G. Bellingham, T.~R. Consi, R.~M. Beaton, and W.~Hall, ``Keeping layered control simple (autonomous underwater vehicles),'' in \textit{Symposium on Autonomous Underwater Vehicle Technology}, June 1990, pp.~3--8.
\bibitem{b2} J.~G. Bellingham \textit{et al.}, ``A second generation survey AUV,'' in \textit{Proc. IEEE Symp. Autonomous Underwater Vehicle Technology}, 1994, pp.~148--155.
\bibitem{b31} E. Potokar, S. Ashford, M. Kaess, and J. G. Mangelson, ``HoloOcean: An underwater robotics simulator,'' in \textit{Proc. IEEE Int. Conf. Robot. Autom. (ICRA)}, May 2022, pp.~3040--3046.
\bibitem{b32} M.~M.~M. Manh\~aes, S.~A. Scherer, M. Voss, L.~R. Douat, and T. Rauschenbach, ``UUV Simulator: A Gazebo-based package for underwater intervention and multi-robot simulation,'' in \textit{Proc. OCEANS 2016 MTS/IEEE Monterey}, Sep. 2016, pp.~1--8.
\bibitem{b34} P. Cieślak, ``Stonefish: An advanced open-source simulation tool designed for marine robotics, with a ROS interface,'' in \textit{Proc. OCEANS 2019 MTS/IEEE Marseille}, 2019, doi: 10.1109/OCEANSE.2019.8867434.
\bibitem{b35} M. M. Zhang \textit{et al.}, ``DAVE Aquatic Virtual Environment: Toward a general underwater robotics simulator,'' in \textit{Proc. IEEE/OES Autonomous Underwater Vehicle Symp.}, 2022, doi: 10.1109/AUV53081.2022.9965808.
\bibitem{b36} D. Davis and D. Brutzman, ``The Autonomous Unmanned Vehicle Workbench: Mission planning, mission rehearsal, and mission replay tool for physics-based X3D visualization,'' in \textit{Proc. 14th Int. Symp. Unmanned Untethered Submersible Technology}, 2005.
\bibitem{b19} F. Liu, H. Tang, Y. Qin, C. Duan, J. Luo, and H. Pu, ``Review on fault diagnosis of unmanned underwater vehicles,'' \textit{Ocean Eng.}, vol.~243, art.~110290, 2022.
\bibitem{b4} B.~Y. Raanan \textit{et al.}, ``Detection of unanticipated faults for autonomous underwater vehicles using online topic models,'' \textit{J. Field Robot.}, vol.~35, no.~5, pp.~705--716, Aug. 2018.
\bibitem{b5} S. Vemprala, R. Bonatti, A. Bucker, and A. Kapoor, ``ChatGPT for robotics: Design principles and model abilities,'' \textit{IEEE Access}, vol.~12, pp.~55682--55696, 2024.
\bibitem{b18} M. Ahn \textit{et al.}, ``Do as I can, not as I say: Grounding language in robotic affordances,'' in \textit{Proc. 6th Conf. Robot Learning (CoRL)}, PMLR vol.~205, 2022, pp.~287--318.
\bibitem{b21} M. Buchholz, I. Carlucho, and Y. R. Petillot, ``A collaborative reasoning framework for anomaly diagnostics in underwater robotics,'' arXiv:2511.03075, 2025 (concurrent preprint).
\bibitem{b16} Y. Bisk, R. Zellers, R. Le Bras, J. Gao, and Y. Choi, ``PIQA: Reasoning about physical commonsense in natural language,'' in \textit{Proc. AAAI}, vol.~34, no.~5, 2020, pp.~7432--7439.
\bibitem{b6} T.~I. Fossen, \textit{Handbook of Marine Craft Hydrodynamics and Motion Control}, 2nd~ed. Chichester, UK: Wiley, 2021.
\bibitem{b11} ggml-org, ``Feature request: support for NVIDIA Nemotron Nano v2,'' llama.cpp, GitHub issue~\#15409, 2025. [Online]. Available: https://github.com/ggml-org/llama.cpp/issues/15409. [Accessed: Jul.~12, 2026].
\bibitem{b14} L. Zheng \textit{et al.}, ``Judging LLM-as-a-judge with MT-bench and Chatbot Arena,'' in \textit{Adv. Neural Inf. Process. Syst. (NeurIPS)}, vol.~36, 2023, pp.~46595--46623.
\end{thebibliography}
\end{document}